\documentclass[lettersize,journal]{IEEEtran}
\usepackage{amsmath,amsfonts}
\usepackage{algorithmic}
\usepackage{array}
\usepackage[caption=false,font=normalsize,labelfont=sf,textfont=sf]{subfig}
\usepackage{textcomp}
\usepackage{stfloats}
\usepackage{url}
\usepackage{verbatim}
\usepackage{graphicx}
\usepackage{subfloat}
\usepackage[graphicx]{realboxes}
\def\BibTeX{{\rm B\kern-.05em{\sc i\kern-.025em b}\kern-.08em
		T\kern-.1667em\lower.7ex\hbox{E}\kern-.125emX}}
\usepackage{balance}

\usepackage{lineno,hyperref}
\usepackage{graphicx}
\usepackage{tabularx}
\usepackage{graphics}
\usepackage{mathrsfs}
\usepackage{amsfonts}
\usepackage{amsmath}
\usepackage{diagbox}
\usepackage{booktabs}
\usepackage[ruled,linesnumbered]{algorithm2e}
\usepackage{caption}
\usepackage{paralist}
\usepackage{lipsum}
\usepackage{amsmath}
\usepackage{amssymb}
\usepackage{indentfirst}
\usepackage{float}
\usepackage{multirow}
\usepackage{hyperref}
\usepackage{multicol}
\usepackage{cite}
\usepackage[misc]{ifsym}
\usepackage[thmmarks,amsmath]{ntheorem}
\usepackage{mathrsfs}
\usepackage{changepage}
\usepackage{bm}
\usepackage{enumitem}
\usepackage{threeparttable}
\usepackage{color}
\usepackage{gensymb}

\usepackage[utf8]{inputenc}
\usepackage{url}
\usepackage{booktabs}
\usepackage{amssymb}
\usepackage{bbding}
\usepackage{pifont}
\usepackage{wasysym}
\usepackage{utfsym}
\usepackage{fontawesome}
\usepackage{makecell}

\usepackage{xcolor}
\usepackage{colortbl}

\begin{document}
	\title{Imbalanced Aircraft Data Anomaly Detection}

	\author{Hao Yang, Junyu Gao, \textit{Member, IEEE}, Yuan Yuan, \textit{Senior Member, IEEE}, Xuelong Li, \textit{Fellow, IEEE}
		\thanks{
			Hao Yang is with the School of Computer Science, and also with the School of Artificial Intelligence, OPtics and ElectroNics (iOPEN), Northwestern Polytechnical University, Xi'an 710072, P. R. China. 
			(E-mail: iopen.yang@mail.nwpu.edu.cn).
			
			Junyu Gao, Yuan Yuan and Xuelong Li are with the School of Artificial Intelligence, OPtics and ElectroNics (iOPEN), Xi'an 710072, P. R. China. They are also with the Key Laboratory of Intelligent Interaction and Applications, Ministry of Industry and Information Technology, Xi'an 710072, P. R. China. (E-mail: gjy3035@gmail.com; y.yuan1.ieee@qq.com; li@nwpu.edu.cn).
			
			\textit{Corresponding author:} Xuelong Li
	}}

	\maketitle
	
	\begin{abstract}
		Anomaly detection in temporal data from sensors under aviation scenarios is a practical but challenging task: 1) long temporal data is difficult to extract contextual information with temporal correlation; 2) the anomalous data are rare in time series, causing normal/abnormal imbalance in anomaly detection, making the detector classification degenerate or even fail. 
		To remedy the aforementioned problems, we propose a Graphical Temporal Data Analysis (GTDA) framework. 
		It consists three modules, named Series-to-Image (S2I), Cluster-based Resampling Approach using Euclidean Distance (CRD) and Variance-Based Loss (VBL).
		Specifically, for better extracts global information in temporal data from sensors, S2I converts the data to curve images to demonstrate abnormalities in data changes. 
		CRD and VBL balance the classification to mitigate the unequal distribution of classes.
		CRD extracts minority samples with similar features to majority samples by clustering and over-samples them.
		And VBL fine-tunes the decision boundary by balancing the fitting degree of the network to each class. 
		Ablation experiments on the Flights dataset indicate the effectiveness of CRD and VBL on precision and recall, respectively.
		Extensive experiments demonstrate the synergistic advantages of CRD and VBL on F1-score on Flights and three other temporal datasets.
	\end{abstract}
	

	\section{Introduction}
	\IEEEPARstart{A}{viation} safety is an important issue in air transportation, receiving much attention from airlines and researches \cite{liu2018robust,wang2019joint,zhou2021safety}.
	The flight test of an aircraft before delivery is an important means of ensuring aviation safety.
	Effective flight tests ensure the aircraft safety and reliability.
	However, when the flight test results are unreliable, the tests can not detect potential problems in aircraft.
	Thus, the sensors in flight tests need periodic calibration \cite{bu2017integrated} to avoid unreliable results in tests.
	It is difficult to get the accurate sensor calibration period directly, and a short calibration period causes huge cost \cite{allerton2005review}.
	Therefore, it makes sense to analyze existing data to judge whether the sensor is abnormal.
	
	\begin{figure}[t]
		\centering
		\subfloat[]{\includegraphics[width=0.12\textwidth]{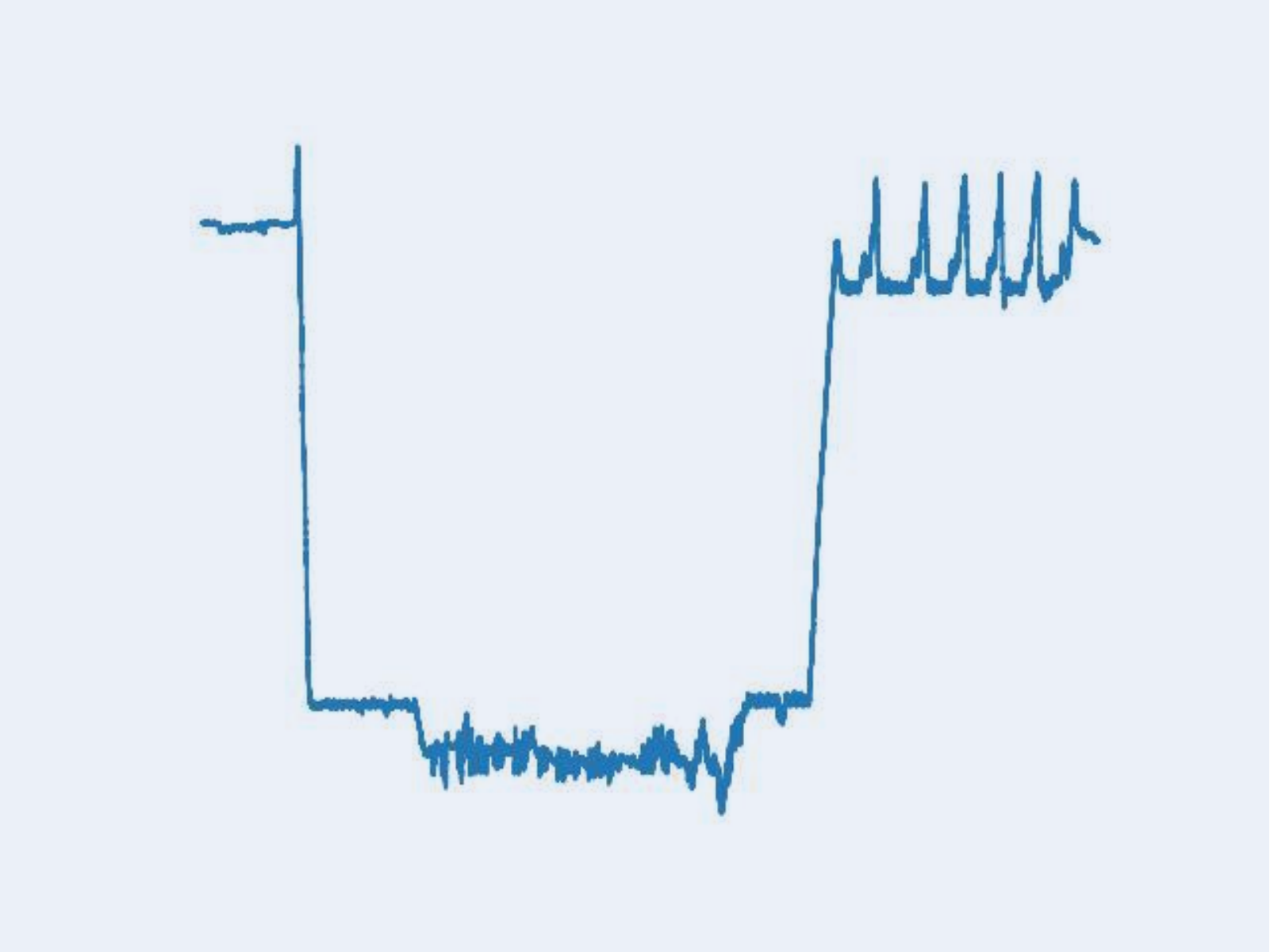}}
		\subfloat[]{\includegraphics[width=0.12\textwidth]{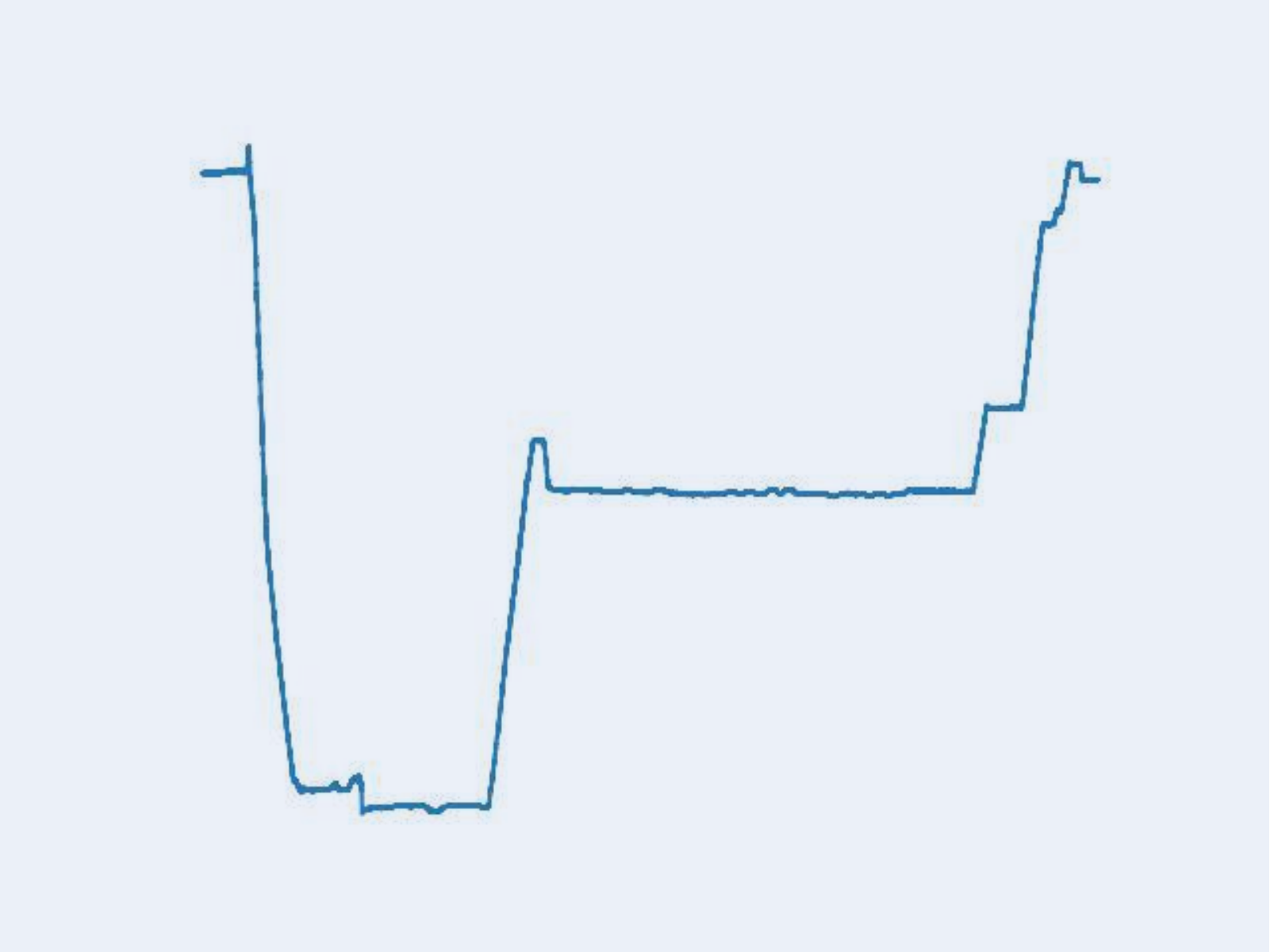}}
		\subfloat[]{\includegraphics[width=0.12\textwidth]{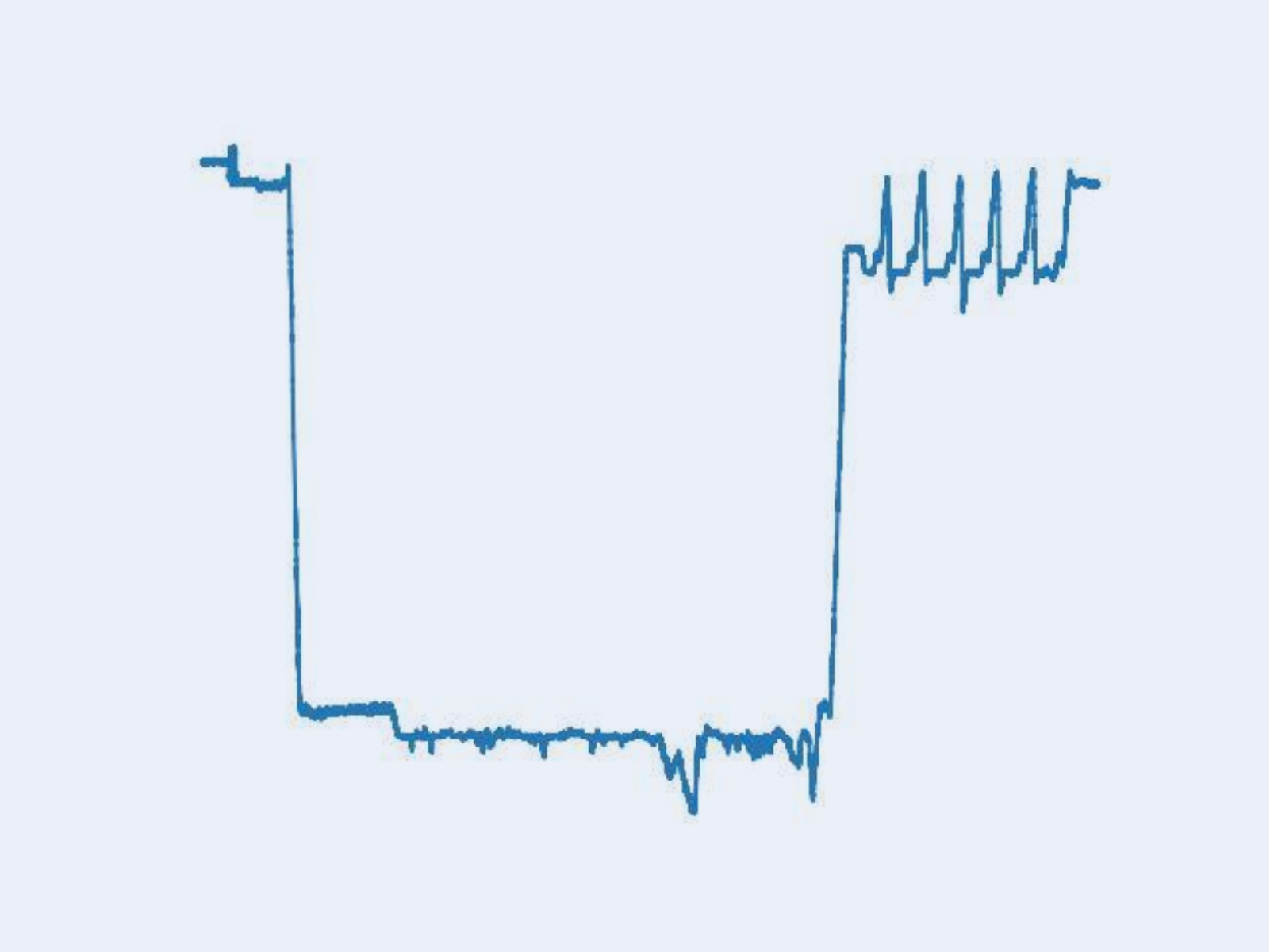}}
		\subfloat[]{\includegraphics[width=0.12\textwidth]{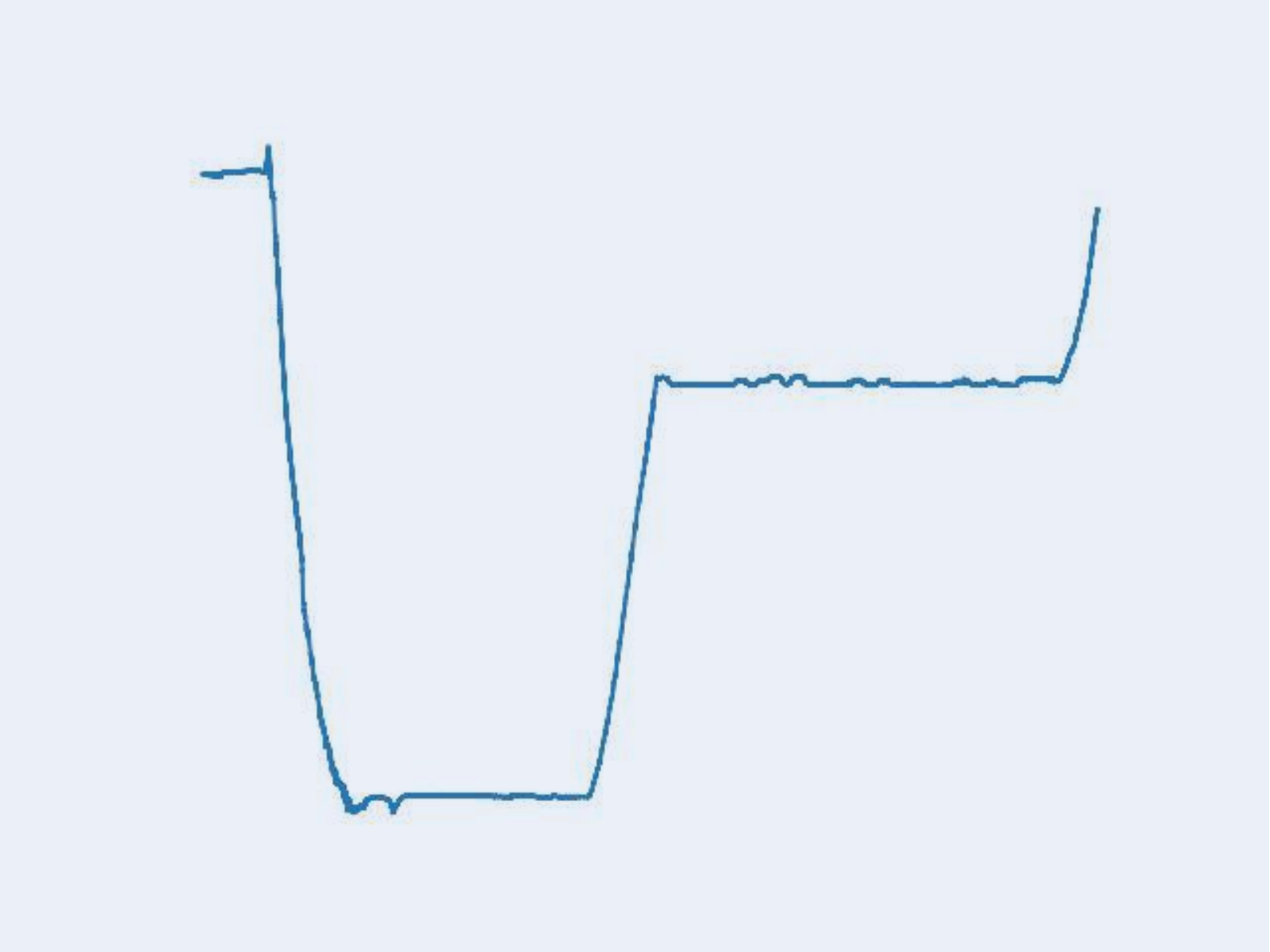}}
		\caption{Four exemplars generated by the imaging technique from the Flights dataset. (a) and (b) are normal data (negative or majority class samples), (c) and (d) are abnormal data (positive or minority class samples).}
		\label{fig1}
	\end{figure}
	
	A sensor anomaly is a degradation of its performance to a certain threshold, usually caused by catastrophic failures and more subtle failures, for instance, outdated calibration, sensor deformation, and low-frequency oscillations \cite{de2020physics}. 
	The performance cannot be directly characterized and is implicit in the flight test data, which is characterized by long time series, imbalance and small differences between classes. 
	Hence, this task is temporal data anomaly detection, suffering from the problems of difficult extraction of global information and scarcity of abnormal data, which makes sensor anomaly detection a challenging task.
	
	For detecting anomaly data accurately, some work attempts to use \textit{Statistical}-based detection methods \cite{tian2021modeling,dietterich1998approximate,goldstein2012histogram}, which need to obtain prior empirical information previously. 
	Nevertheless, in complex aviation scenarios, the data distribution often can not be represented using conventional distributions.
	Similarly, \textit{traditional machine learning}-based methods \cite{iiduka2021appropriate,hosseinzadeh2021improving,perdisci2006using,saraswat2021ansmart,nie2020decision}, extract manual features from data, but sensor anomaly information is implied in the fight test data, making it difficult to determine and extract effective features.
	At the same time, the above two methods are also insensitive to time series.
	\textit{Natural Language Processing}-based methods \cite{park2018multimodal} have difficulty in covering global contextual information when encountering ultra-long time-series data, which also leads to these methods not working in this task.
	\textit{Computer Vision} (CV)-based anomaly detection methods \cite{Deng_2022_CVPR, lin2021learning, gao2021audio} are well developed for anomaly detection of images. 
	However, due to the difference in data formats, CV-based methods are challenging to operate directly in temporal data anomaly detection.
	
	Inspired by the speech recognition field, the sequence data are convert to images.
	For example, speech voice data are often pre-processed by MFCCs or PLPs \cite{hermansky1990perceptual} before being fed into the network. 
	Some works \cite{wang2015imaging,marwan2002recurrence} attempt to convert temporal data into images and use CV-based methods to solve the temporal data anomaly detection problem.
	Examples include Recurrence Plot (RP) \cite{marwan2002recurrence}, Markov Transition Fields (MTFs) \cite{wang2015imaging}, etc. However, the images generated by RP suffer from the problem of ambiguity with the original time series. 
	The size of an image generated by Gramian Angular Fields (GAFs) \cite{wang2015imaging} is positively correlated with the length squared term of the temporal data.
	For this particular task, i.e. temporal data of length approximately $ 250,000 $, the image generated by GAFs is extremely memory intensive and difficult to train for the network, as general convolutional kernels do not cover such large images.
	
	In addition to the problems raised above, the imbalanced data distribution is also particularly important in anomaly detection, which leads to significant performance degradation of classical classification network architectures in imbalanced datasets. 
	Specifically, the model is more inclined to learn majority class features and ignore minority class features, which results in the network predicting all samples as majority class during training. It is also easy to rely on existing data samples and over-fitting problems for trained models.
	
	To remedy the abovementioned problems,  a Graphical Temporal Data Analysis (GTDA) framework is developed in this paper to tackle anomaly detection in temporal data.
	It can not only convert the original one-dimensional data format into an image format based on maintaining the data time series relationship but also control the size of the generated image.
	Besides, it can also change the data distribution by oversampling to eliminate its influence.
	Specifically, the framework is divided into two steps:
	
	1) Primarily, we propose the \textbf{S}eries-\textbf{to}-\textbf{I}mage (S2I) method.  
	A time-valued rectilinear coordinate system is established to reflect the temporal data features directly from the image.
	Moreover, the proposed method can control the size of image by controlling the range of the horizontal and vertical coordinates, avoiding the problem of difficult training of the network caused by too large generated images.
	2) Additionally, to alleviate the problem of uneven data distribution, we propose \textbf{C}luster-based \textbf{R}esampling approach using Euclidean \textbf{D}istance (CRD) and \textbf{V}ariance-\textbf{B}ased \textbf{L}oss (VBL).
	CRD identifies samples with similar characteristics between two classes by clustering, and then over-samples these samples to enable the network to better distinguish these data. 
	At the same time, the Variance-Based Loss method is proposed to fine-tune the decision boundary as well as to stabilize the training.
	We argue that the network has worse aggregation ability for the class with higher variance (the distribution of values calculated by the softmax function after network inference). Therefore, greater weights are given to such class.
	
	In summary, our main contributions are:
	
	\begin{itemize}
		
		
		
		\item{Develop a GTDA framework for ultra-long temporal data anomaly detection. The input temporal data are first convert to image by a Series-to-Image (S2I) module, and then a CNN-based model extracts the features and judges whether the anomaly occurs. The proposed GTDA framework allows CV-based method to process long-series contextual information directly.}
		
		\item{Design a Cluster-based Resampling approach using Euclidean Distance (CRD) method to oversample the minority samples. Experiments show that CRD can enhance the precision of the model.}
		
		\item{Present a Variance-Based Loss (VBL) to balance the classification of each class by adjusting the loss adaptively. Experiments show that VBL can promote the recall of the model.}
		
	\end{itemize}
	
	\section{Related Work}
	
	\subsection{Anomaly Detection}
	The mainstream flight-related anomaly detection methods are currently clustering-based \cite{marcos2013clustering,pu2020hybrid,kiss2014data,li2021clustering}, neighborhood-based \cite{ghosh2016anomaly,breunig2000lof,ester1996density}, regression-based \cite{liu2016regression,cheng2015gaussian,oh2018residual} and classification-based \cite{bergman2020classification,ruff2018deep,golan2018deep,han2020unsupervised} methods. 
	Clustering-based \cite{marcos2013clustering,pu2020hybrid,kiss2014data,li2021clustering} and neighborhood-based \cite{ghosh2016anomaly,breunig2000lof,ester1996density} methods require mining and exploiting relationships between data and require a high level of expert domain knowledge. 
	Regression-based methods \cite{liu2016regression,cheng2015gaussian,oh2018residual} detect anomalies by fitting to serial data and detecting anomalies ground on the residuals between inferred and actual values.
	But the regression-based method is difficult to adapt to the dynamically changing parameters of the aircraft during operation. Classification models \cite{bergman2020classification,ruff2018deep,golan2018deep,han2020unsupervised,wang2020looking,nie2019multiview} can also perform anomaly detection. 
	Nonetheless, the model performance significantly degrades when an imbalanced dataset is involved. 
	Recently, Numerous studies \cite{prusa2015using,yen2009cluster,yen2006under,morik1999combining,xie1989logit,cao2019learning,menon2020long,tan2020equalization,wu2020forest} attempted to alleviate the mentioned problem by designing balanced sampling methods or balanced loss functions, such as oversampling or undersampling the data. 
	
	\subsection{Imaging Temporal Data}
	Encouraged by the great success of CV in classification tasks, some works attempt to visualize temporal data and then use the network architecture of CV to solve temporal tasks.
	In speech recognition systems, speech signals from one-dimensional data are converted to two-dimensional data using  MFCCs or PLPs \cite{hermansky1990perceptual}.
	RP \cite{eckmann1995recurrence} can reveal the internal structure of time series and analyse temporal periodicity, non-smoothness, etc.
	Short Time Fourier Transform \cite{parchami2016recent} splits the time domain signal into multiple segments through a sliding window and then performs an fast Fourier transform (FFT) on each segment to generate the time-frequency spectral information of the signal.
	GAFs \cite{wang2015imaging} convert the temporal information from a right angle coordinate system to a polar coordinate system, using angles or angular differences to represent the time domain information.
	The time-insensitive Markovian transfer matrix is taken into account for temporal information to propose MTFs \cite{wang2015imaging}.
	Unlike the above methods, the proposed module is more flexible and practical in imaging ultra-long temporal data and can better combine the advantages of classification models on intuitive images.

	\begin{figure*}[t]
		\centering
		\includegraphics[width=1.0\textwidth]{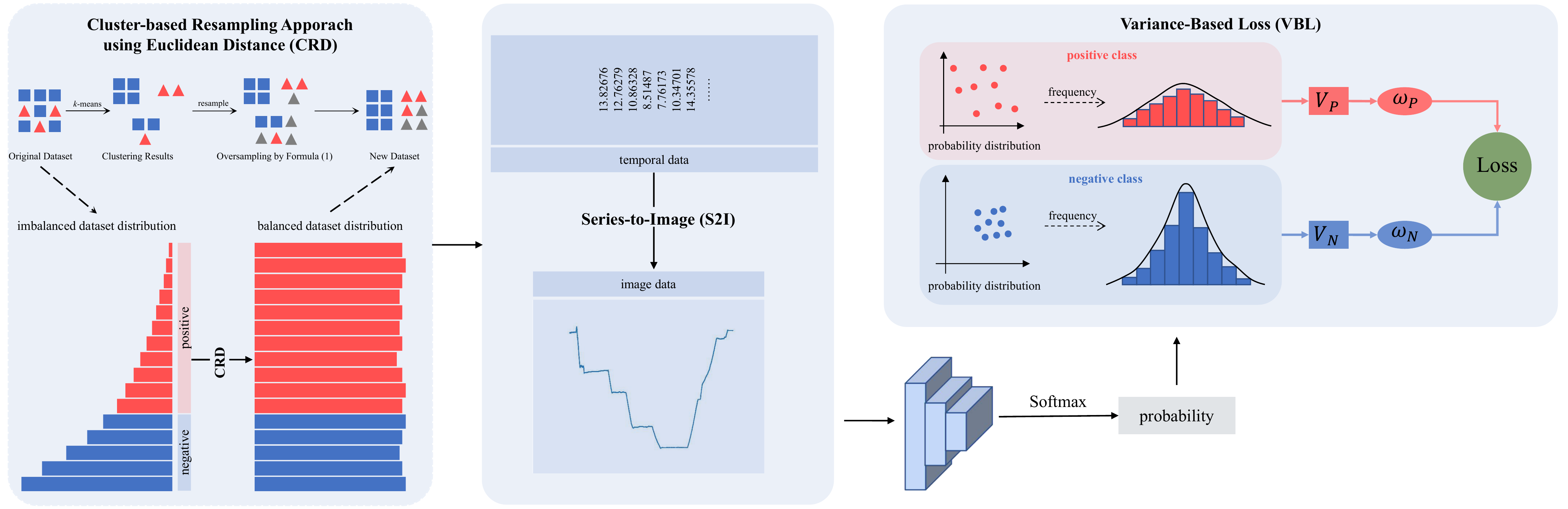}
		\caption{The flowchart of the proposed GTDA framework, which contains three modules: 1) CRD is designed to balance datasets by changing the data distribution. 2) S2I is used to imaging the temporal data. 3) VBL weights the loss function based on the data distribution after the classifier model inference.}
		\label{fig2}
	\end{figure*}
	
	\subsection{Imbalanced Learning}
	Imbalanced learning alleviates the problem of model performance degradation on imbalanced datasets from two perspectives: re-sampling and re-weighting.
	Resampling methods focus on adding or subtracting samples from the training set.
	The random undersampling approach (RUS) \cite{prusa2015using} discards the majority class samples to equilibrate the loss.
	The SBC approach \cite{yen2009cluster,yen2006under} determines the majority number per cluster by clustering the samples of each cluster's majority class by selecting them in a different way. 
	SMOTE \cite{chawla2002smote} is established on the KNN to manually synthesize the majority class examples.
	The Borderline-SOMTE \cite{han2005borderline} algorithm solves the problem of sample overlap in the sample generation process of the SMOTE algorithm. 
	There is also some works on generating pseudo-minority class samples based on VAE \cite{fabius2014variational,an2015variational,khalid2020oc,zhou2021vae} and GAN \cite{kim2020gan,zenati2018efficient,jiang2019gan}.
	Reweighting methods focus on optimizing the weights of different classes.
	Most works \cite{morik1999combining,xie1989logit,cao2019learning,menon2020long,tan2020equalization,wu2020forest} change the weights according to the sample number per class.
	The weights of loss functions in diverse classes designed by \cite{morik1999combining,xie1989logit} are in inverse ratio to the corresponding sample number.
	Different from existing approach, we calculate the weights for each class adaptively by computing the variance of the samples inferred from the network. It can dynamically express the degree of aggregation of the network for each class of samples.

	\section{Methodology}
	\subsection{Overview}
	In this section, we develop a Graphical Temporal Data Analysis (GTDA) framework. 
	First, a graphical approach, named S2I (Section III.B), is proposed to convert one-dimensional temporal data into images.
	Then, a resampling method CRD (Section III.C) is designed to find suitable minority class samples for oversampling by clustering, to alleviate the above problems and to be able to achieve coarse tuning of decision boundary.
	In addition to this, we fine-tune the bounds by VBL (Section III.D), which characterizes the degree of aggregation of the model by variance for each category. 
	Fig. \ref{fig2} illustrate the overall framework.
	
	\subsection{Series-to-Image (S2I)}
	
	In this section, we propose \textbf{S}eries-\textbf{to}-\textbf{I}mage (S2I) to transform one-dimensional temporal data into images for feeding into a general architecture for CV. 
	This is because in practical scenarios, such as during flight test, anomalies span a large time series and are difficult to be handled by NLP-based methods such as RNN and LSTM. 
	Therefore, a framework \ref{fig2} is proposed that allows the model to encode the temporal data in the same way as processing images and flexibly introduces various CV-based classification and anomaly detection modules to improve the model's capability to model semantic or contextual knowledge on a large scale range. 
	Besides, we want to take advantage of the CV-based classification methods on intuitive images and try to understand the abnormalities implicitly embedded in the data changes from the waveform graph perspective in order to visualize the value shifts from the time dimension.. 
	To this end, a waveform graph criterion called S2I is designed for temporal data to image. 
	We explore the effectiveness of the image-based data approach through CV techniques, starting from several major parameters that affect waveform graphs. This approach has two advantages: 1) Fully exploiting the role of CV-based modules in advanced semantic parsing and global semantic coverage, and 2) Effectively using classification tasks and anomaly detection methods in CV.

	\subsection{Cluster-based Resampling Approach using Euclidean Distance (CRD)}
	Due to the extreme imbalanced distribution between classes, the \textbf{C}luster-based \textbf{R}esampling approach using Euclidean \textbf{D}istance (CRD) method is proposed to alleviate the imbalance in the training datasets distribution by oversampling the minority class samples.
	We first cluster all the training samples into some clusters. 
	According to \cite{yen2009cluster}, the samples in different clusters have different features and they have similar features when they are in the one cluster. 
	So if there are more majority classes in a cluster, that cluster has more majority class features. 
	However, unlike \cite{yen2009cluster }, which retains more prominent features when under-sampling, we believe that such clusters, which with far more negative samples than positive, or conversely, already have distinct negative or positive sample features, and therefore do not need to retain more information when over-sampling or under-sampling.
	With this in mind, we propose CRD.

	\begin{figure}[t]
		\centering
		\includegraphics[width=0.45\textwidth]{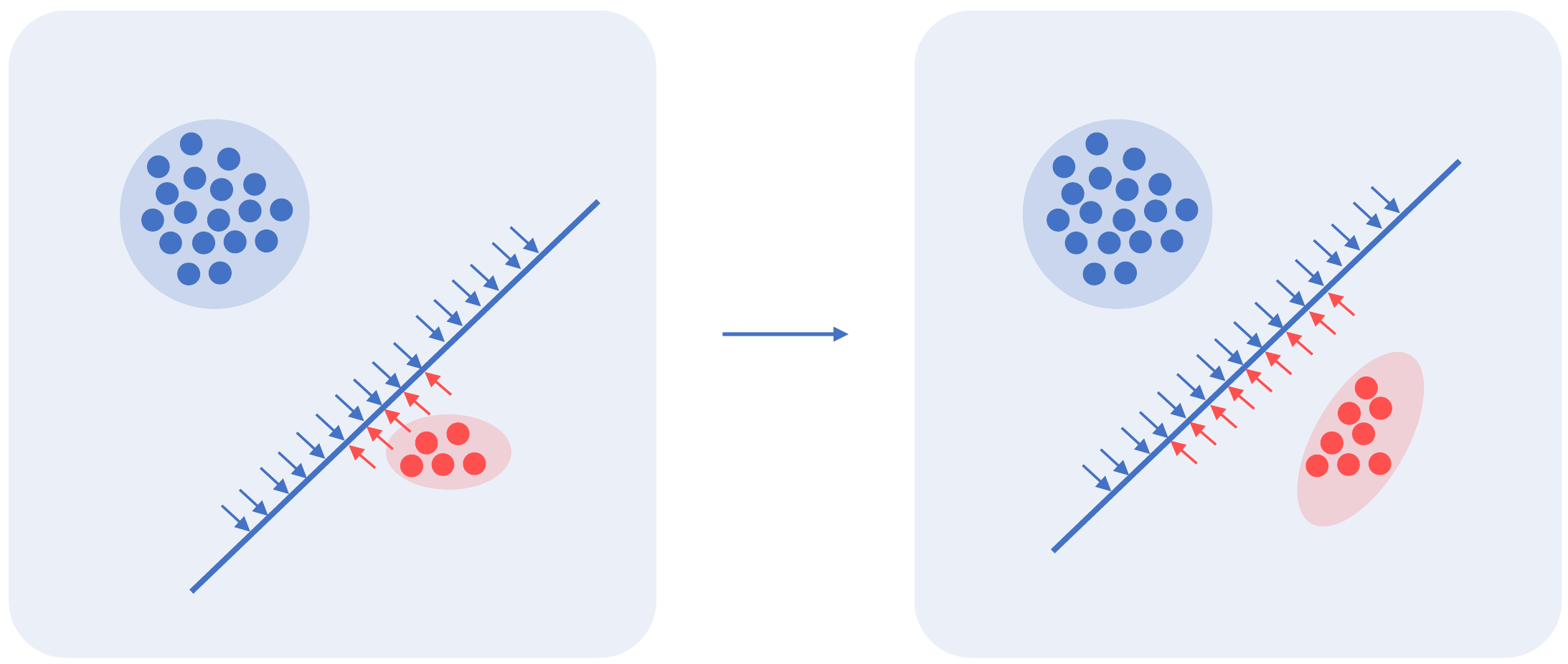}
		\caption{CRD. Oversample positive samples close to the decision boundary to improve the model’s ability to discriminate similar samples from different classes.} 
		\label{fig3}
	\end{figure}
	
	\renewcommand{\arraystretch}{1.3} 

	Let the number of samples in the training set be $N$, where the number of majority (negative) samples is $N_{MA}$ and the number of minority (positive) samples is $N_{MI}$. we first cluster the training set into $ k $ clusters.
	$ N_{MA}^i $ and $ N_{MI}^i $ denote the number of negative and positive class samples in the $ i^{th} $ cluster, respectively. Therefore, the ratio of negative class to positive in the $ i^{th} $ cluster is $ {N_{MA}^i}/{N_{MI}^i} $. 
	Assume that the ratio of the negative and positive sample number after generation is set to $ m:1 $, then the formula is as follows:

	\begin{equation}
		\label{CRD}
		\begin{aligned}
			N\!N_{MI}^i = \frac{N_{MA}^i}{m} \times \frac{1}{k-1} \times \sum_{i=1}^{k} (1 - \frac{N_{MA}^i / N_{MI}^i}{\sum_{j=1}^{k} {N_{MA}^j / N_{MI}^j}} ) ,
		\end{aligned}
	\end{equation}
	
	\noindent where $ \frac{N_{MA}^i}{m} $ indicates the sample number in the positive class in the training set after oversampling, and $ {\sum_{j=1}^{k} {N_{MA}^j / N_{MI}^j}} $ indicates the total proportion of majority and minority classes in each cluster in the original training set.
	Clusters with larger proportions, i.e. clusters with more majority or minority class features, are easier to classify during training.
	Clusters with smaller proportions indicate that the positive and negative samples have similar features, and also indicate that these samples are close to the decision boundary. 
	By oversampling the positive examples of these clusters, the network is more likely to extract their features and distinguish them..
	Therefore, $ 1 - \frac{N_{MA}^i / N_{MI}^i}{\sum_{j=1}^{k} {N_{MA}^j / N_{MI}^j}}  $ is designed to reduce the oversampling frequency of the proportionally larger clusters and increase the oversampling frequency of the positive samples in the proportionally smaller clusters.
	The value of $ \sum_{i=1}^{k} (1 - \frac{N_{MA}^i / N_{MI}^i}{\sum_{j=1}^{k} {N_{MA}^j / N_{MI}^j}} )  $ can be calculated to be $ k-1 $ and therefore multiplied by $ \frac{1}{k-1} $, making the whole coefficient considered as a weighting.
	
	In the process of clustering, the number of samples in each cluster can be determined by using formula \ref{CRD}.
	And in the concrete implementation, the Euclidean distance is used to calculation the distance between any two samples.

	\begin{table}[h]
		\centering
		\fontsize{8}{10}\selectfont
		\caption{Functions or variables in Alg.1 and Alg.2.}
		\begin{tabular}{cc}
			\Xhline{1.2pt}
			\makebox[0.16\textwidth][c]{Functions\&Variables} 
			& \makebox[0.22\textwidth][c]{Explanation}  \\
			\Xhline{1.2pt}
			$ X' $ & original training dataset  \\
			$ y' $ & original ground truth label  \\
			$ X $ & training dataset after CRD  \\
			$ y $ & ground truth label after CRD  \\
			$ u $ & cluster label \\
			$ k\!-\!means $ & k-means algorithm \\
			$ cal\_1 $ & calculate the number of samples per cluster \\
			$ cal\_2 $ & calculate $ N\!N_{MI} $ by ( \ref{CRD} )\\
			$ ROS $ & Random Over-Sampling \cite{prusa2015using} operation \\
			\Xhline{0.8pt} 
			$ n $ & recursion times\\
			$ P $ & probability\\
			$ S() $ & softmax operation \\
			$ \omega\_convert $ & convert $\omega$ by ( \ref{convert} ) \\
			$ \omega\_update $ & update $ \omega $ by ( \ref{weight} ) \\
			$ variance\_update $ & update $ V $ by ( \ref{VandA1} ) \\
			$ mean\_update $ & update $ A $ by ( \ref{VandA2} ) \\
			$ \alpha\_update $ & update $ \alpha $ by ( \ref{VandA3} ) \\
			$ loss\_compute $ & compute $ loss $ by ( \ref{VBL} ) \\
			\Xhline{1.2pt} 
		\end{tabular}
		\label{table1}
	\end{table}

	\subsection{Variance-Based Loss (VBL)}
	We then propose a \textbf{V}ariance-\textbf{B}ased \textbf{L}oss (VBL) to equilibrate the loss during training by adaptively attaching weights to each class.
	The design of VBL is considered from two aspects: 
	1) In the object classification, cross-entropy loss is a common and effective function for balanced datasets.
	But in imbalanced datasets, the small number of abnormal samples and the equiprobability of sample selection in the training process make the minority class contribute less to the loss function, which eventually leads the model to tend to predict all samples to be predicted as majority class.
	2) Although CRD pulls the decision boundary back between the two classes and moves it away from the minority class by oversampling the minority class more, the decision boundary is still sometimes difficult to train. 
	This is because the redundancy of the data generated during the execution of CRD still causes the model to be easily over-fitted to the minority class. 
	
	\begin{algorithm}[t]
		\caption{Algorithm to realize CRD.}
		\label{alg1}
		\KwIn{$ (X', y') $, $ m $, and $ k $}
		\KwOut{$(X, y)$}
		\textbf{Initialization}: $ m \leftarrow 1 $, $ k \leftarrow 6 $ \\
		\textit{\# Cluster the $ X' $ into $k$ clusters.} \\
		Cluster $ (X', u) \leftarrow k\!-\!means(X', k) $; \\
		\textit{\# Calculate $ N_{MA} $ and $ N_{MI} $ in each cluster.} \\
		Calculate $ N_{MA}, N_{MI} \leftarrow cal\_1(X', y', u) $; \\
		\textit{\# Calculate $ N\!N_{MI} $ in each cluster after CRD.} \\
		Calculate $ N\!N_{MI} \leftarrow cal\_2(N_{MA}, N_{MI}, k) $; \\
		\textit{\# Generate the new balanced training dataset.} \\
		Generate $ (X, y) \leftarrow ROS(X', y', u, N\!N_{MI}) $; \\
		\textbf{return} $(X, y)$. \\
	\end{algorithm}
	
	To stabilize the training process, some recent works \cite{li2020overcoming,ren2020balanced,tan2020equalization,wang2021seesaw} add weights to the corresponding losses by the number of samples from different classes.
	However, as mentioned in Section II.D, the sample number is hardly effective to characterize the degree of training of the samples in the training process. Unlike previous approaches, VBL characterizes the degree of aggregation of the network for each class by the variance inferred from the model.
	Specifically, if a class has a large variance which is inferred by the model, it means that the network is not aggregating this class well enough, and we want VBL to give more weight to this class so that the decision boundary is far from this class and close to the other class.
	To implement this idea, we add variance-related adaptive weights to the loss during the training process.
	
	Formally, let $ F $ represents a mapping from an input sample $ x $ to the prediction after model inference, and generates a class vector $ z=F(x) $, where $ z \in \mathbb{R}^2 $ (positive or negative class). $ y $ indicates the corresponding ground  truth label and $ y \in \{ \cal{P}, \cal{N} \} $. Where $ \cal{P} $ and $ \cal{N} $ denote positive and negative class, respectively. 
	VBL can be expressed by formula \ref{VBL}. 
	
	\begin{equation}
		\label{VBL}
		\begin{aligned}
			{\cal{L}}(z, y)	&= -\log \frac{\widetilde{\omega}_{y} e^{z_y}}{\sum_{\hat{y} \in \{\cal{P,N}\}} {\widetilde{\omega}_{\hat{y}}} e^{z_{\hat{y}}}}  \\
			&=\log \left( \sum_{\hat{y} \in \{\cal{P,N}\}} \frac{\widetilde{\omega}_{\hat{y}}}{\widetilde{\omega}_{y}} e^{z_{\hat{y}} - z_{y}} \right), 
		\end{aligned}
	\end{equation}
	
	\begin{figure}[t]
		\centering
		\includegraphics[width=0.45\textwidth]{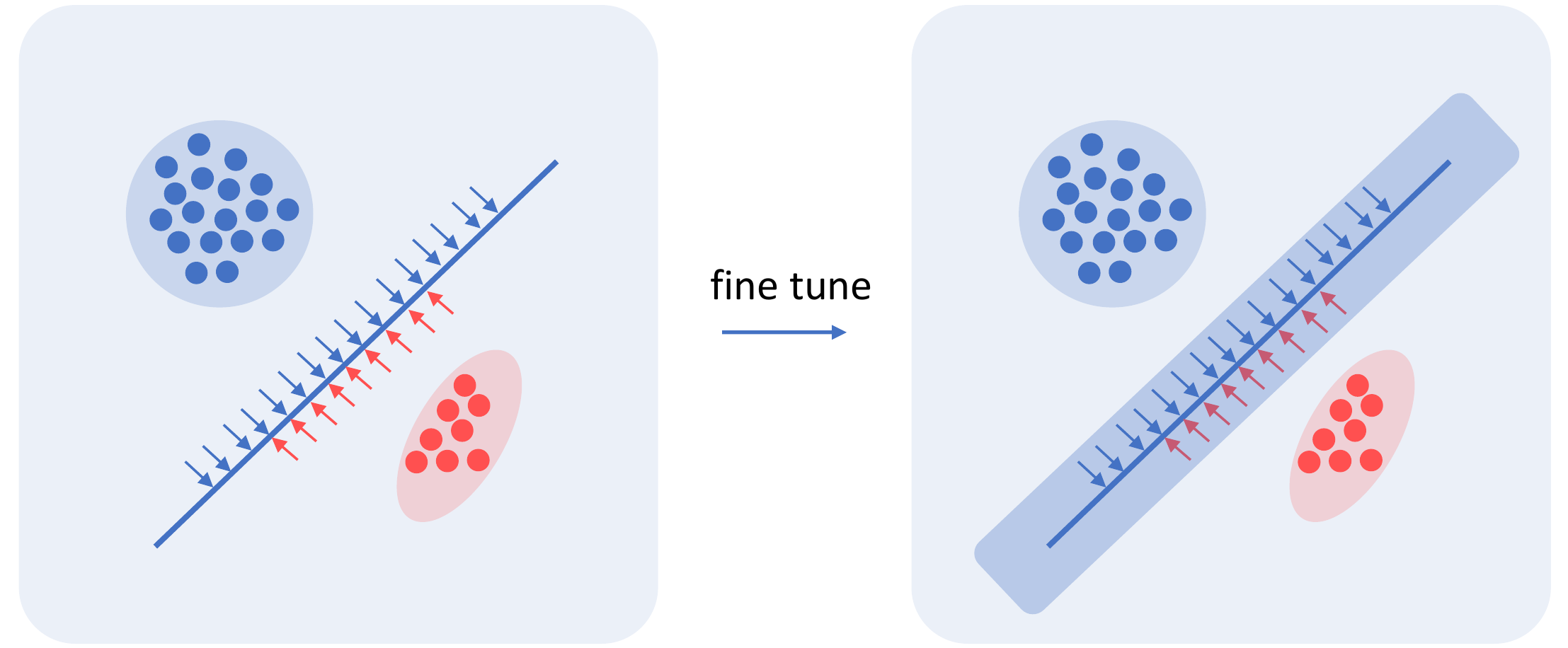}
		\caption{VBL. Modifying the intensity of the decision boundary adjustment adaptively according to the fitting degree of the model to each class.} 
		\label{fig4}
	\end{figure}

	\noindent where $ \widetilde{\omega} $ represents the weight corresponding to the ground truth label. Weights are variance-dependent, and the computation of weights is proportional to the variance. 
	And in the previous analysis, the loss is in inverse ratio to the variance, so the loss function needs to be inversely proportional to the weights. 
	On this basis, the representation of the weight can be given by formula \ref{convert}.
	
	\begin{equation}
		\begin{aligned}
			{\widetilde{\omega}_{y}} &= \begin{cases}
				\omega_{\cal{P}}, & y = \cal{N} \\
				\omega_{\cal{N}}, & y = \cal{P}
			\end{cases}. \\	
		\end{aligned}
		\label{convert}
	\end{equation}
	
	\begin{algorithm}[t]
		\caption{Algorithm to realize VBL.}
		\label{alg2}
		\KwIn{label $y$, and outputs $z$}
		\KwOut{\bm{$loss$}}
		
		\textbf{Initialization}: \\
		$\gamma \leftarrow \frac{batchsize}{N},
		A \leftarrow 0, 
		V \leftarrow 0, $ \\
		$ n \leftarrow 0,
		\alpha \leftarrow 1, 
		\omega \leftarrow [1, 1] $.\\
		\eIf{$ \alpha > 0 $}
		{
			Calculate $ P \leftarrow S(z) $; \\
			Update $n \leftarrow n + 1$; \\
			\eIf{$ n == 1 $}
			{Update $ A \leftarrow P $;}
			{Update $ A \leftarrow mean\_update(A, P, n) $;}
			Update $ V \leftarrow variance\_update(V, A, P, n) $; \\
			Update $ \omega \leftarrow \omega\_update(\omega, \alpha, V) $; \\
			Convert $ \omega \leftarrow \omega\_convert(\omega) $; \\
			Update $\alpha \leftarrow \alpha\_update(\gamma) $; \\
		}
		{
			\textit{\# No operation in else-block.}
		}
		Compute $ loss \leftarrow loss\_compute(\omega, z) $;\\
		\textbf{return} \bm{$loss$}. \\
	\end{algorithm}

	In Equation \ref{VBL}, $ {\cal{L}}(z, y) $ adaptively adjusts the weights of each class based on the variance generated by each class during the network training. 
	The aim is to have more weight for classes with poor model aggregation. That is, the class with larger variance has more weight and the class with better aggregation has less weight. 
	$ \omega_{y} $ can be expressed by formula \ref{weight}.
	\begin{equation}
		\label{weight}
		\begin{aligned}
			& \omega_{y,n} = \alpha_n \omega_{y,n-1} + (1 - \alpha_n) V_{y,n}, \\
		\end{aligned}
	\end{equation}
	
	\noindent where $ \omega_{y} $ is given using the recursive formula and $ n $ denotes recursion times. This is because the variance $ V_{y} $ needs to be accumulated during the training process. According to the law of large numbers, enough samples are needed to fit the overall distribution, so $ \omega_{y} $ cannot be directly replaced by $ V_{y} $. $ \alpha $ is the linear decay factor. 
	The calculation of $ V_{y} $ requires the mean value $ A_{y} $. $ V_{y, n} $, $ A_{y, n} $ and $ \alpha_n $ are given by: 
	
	\begin{equation}
		\label{VandA1}
		V_{y,n} = \frac{n-1}{n^{2}}\left(S(z_y)-A_{y, n-1}\right)^{2}+\frac{n-1}{n} V_{y,n-1}, 
		\vspace{-0.2cm}
	\end{equation}
	
	\begin{equation}
		\label{VandA2}
		A_{y, n} = A_{y, n-1}+\frac{S(z_y)-A_{y, n-1}}{n},
		\vspace{-0.30cm}
	\end{equation}
	
	\begin{equation}
		\label{VandA3}
		\alpha_n = \alpha_{n-1} - \gamma,
	\end{equation}
	
	\noindent $ S(z_y) $ in Equation \ref{VandA2} denotes Softmax, which converts the model outputs $ z_y $ into the prediction probability distribution. $ \gamma $ in Equation \ref{VandA3} is a constant hyperparameter.

	\section{Experiments}
	This section is divided into four subsections, which include experimental setup, ablation experiments, experimental results and experimental analysis. In the experimental setup, we introduce the evaluation metrics, the data set and the implementation details. 
	A full-scale ablation experiment then validates the efficacy of the proposed modules. 
	The experimental results of the GTDA are presented in the four selected datasets.
	Finally, in the experimental analysis subsection, this paper analyzes the effectiveness of GTDA based on extensive experimental comparisons
	
	\begin{figure*}[t]
		\centering
		\subfloat[]{\includegraphics[width=0.4\textwidth]{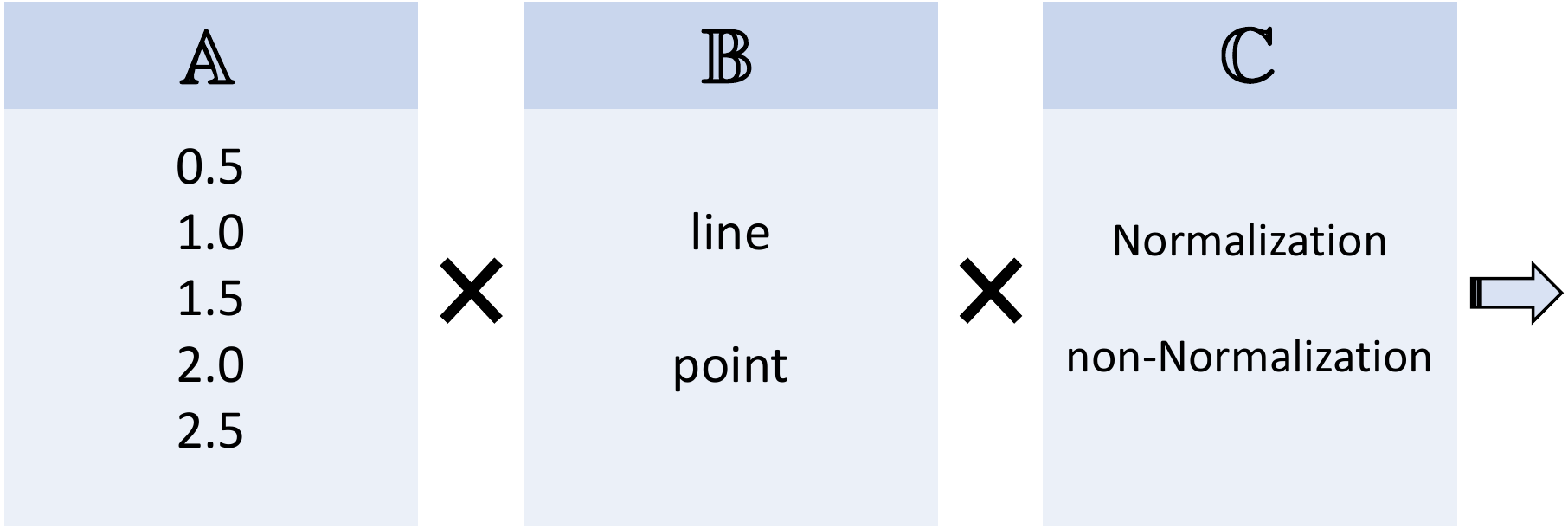}}
		\subfloat[]{\includegraphics[width=0.2\textwidth]{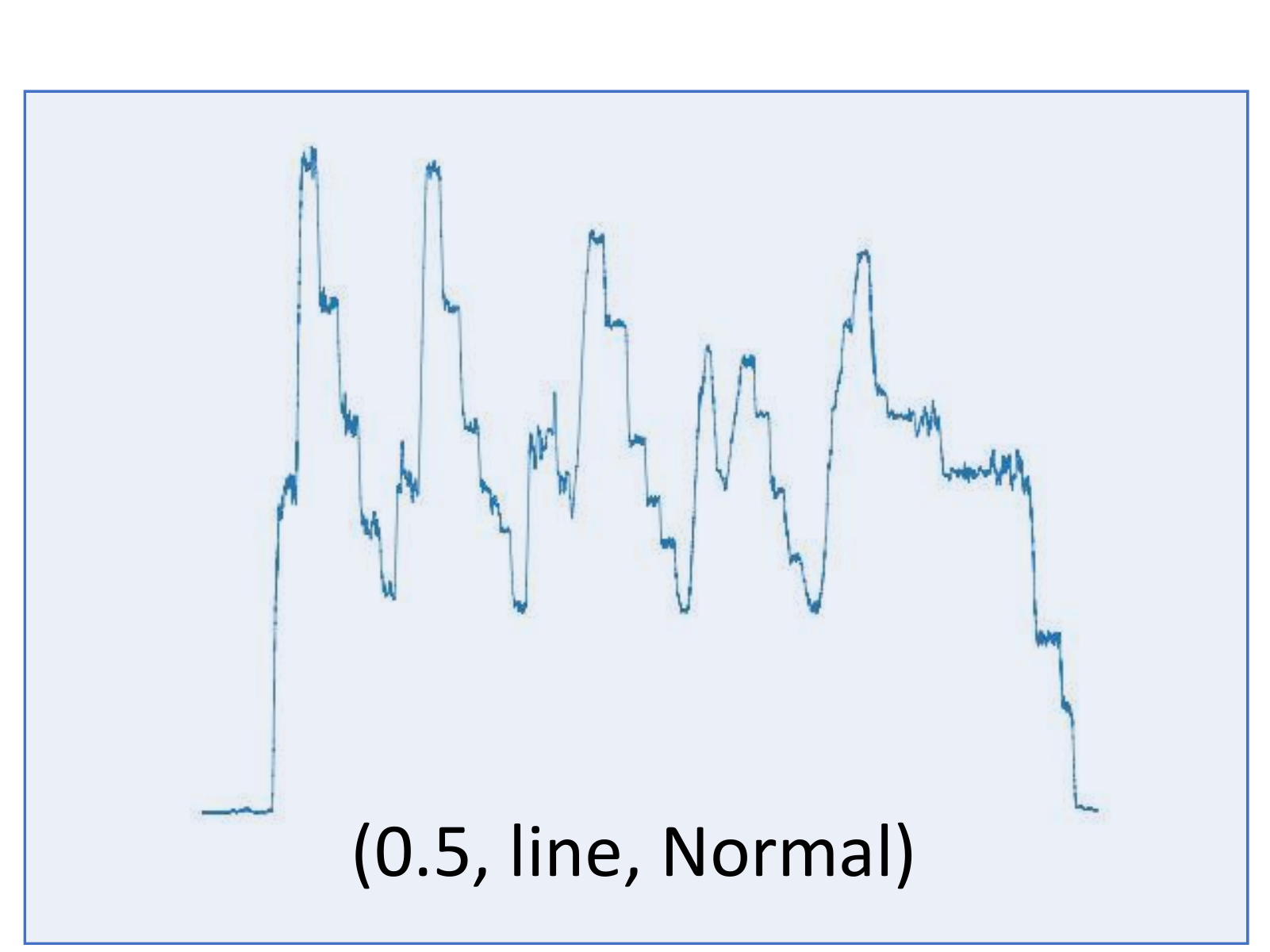}}
		\subfloat[]{\includegraphics[width=0.2\textwidth]{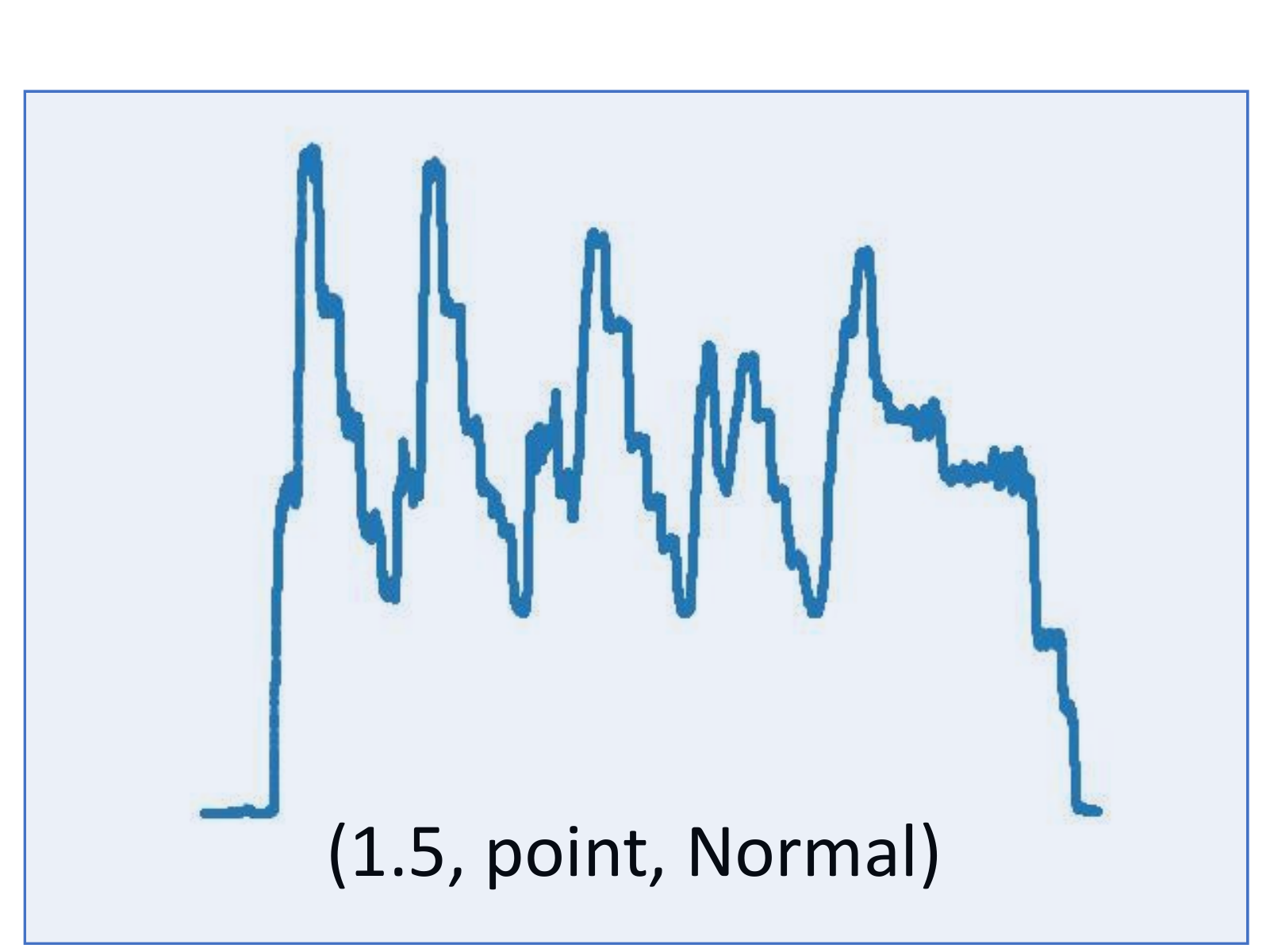}}
		\subfloat[]{\includegraphics[width=0.2\textwidth]{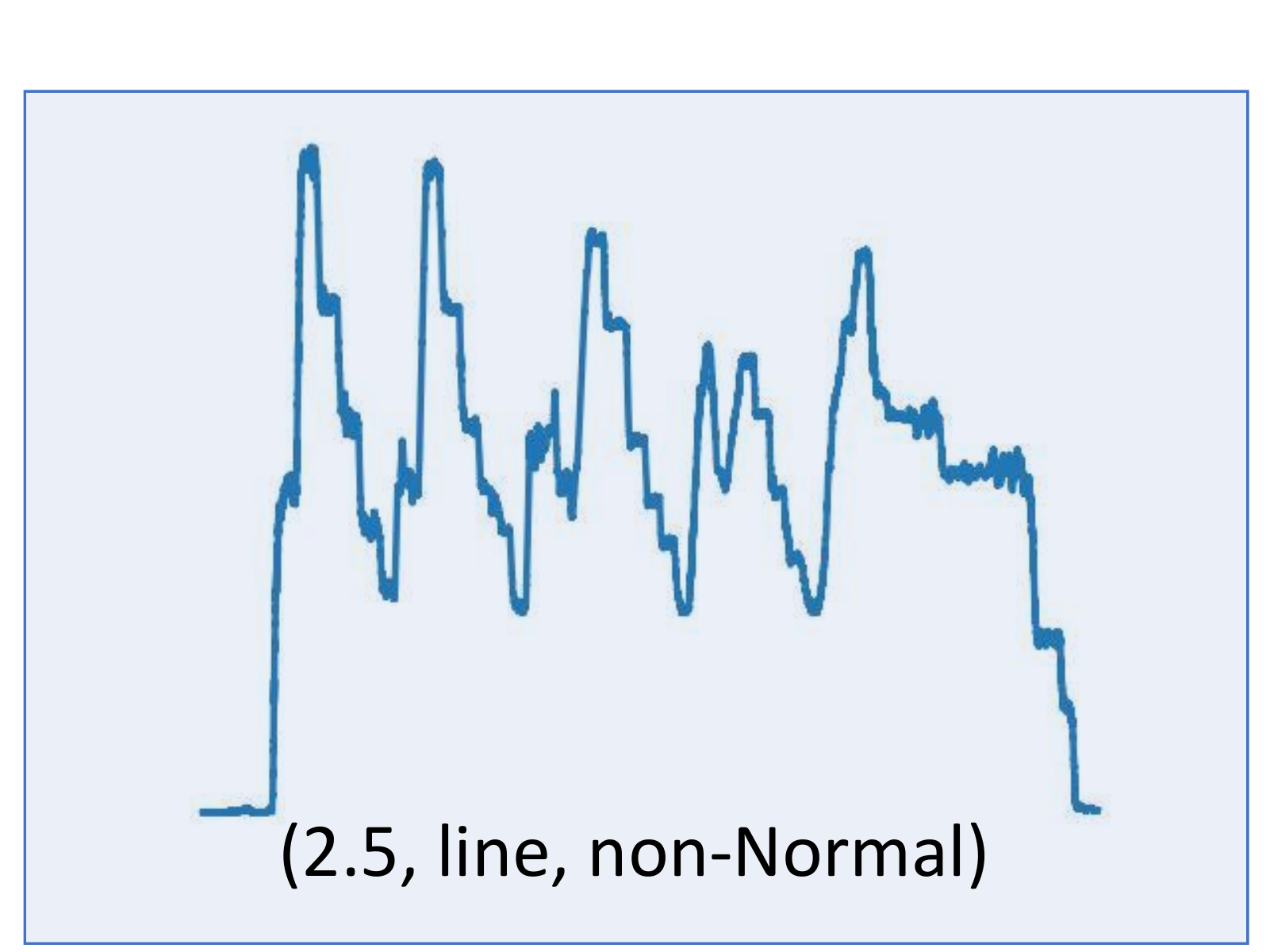}}
		
		\caption{Setting of S2I parameters and images generated under specific parameters. (a) The process of Cartesian product about the curve scale $ \mathbb{A} $, type $ \mathbb{B} $ and the normalization operation $ \mathbb{C} $. $ \times $ indicates Cartesian Product. $ (0.5, \text{line}, \text{Normal}) $, $ (1.5, \text{point}, \text{Normal}) $ and $ (2.5, \text{line}, \text{non-Normal}) $ in (b), (c) and (d), respectively.}
		\label{fig5}
		
	\end{figure*}

	\begin{figure*}[t]
		\centering
		\subfloat[]{\includegraphics[width=0.25\textwidth]{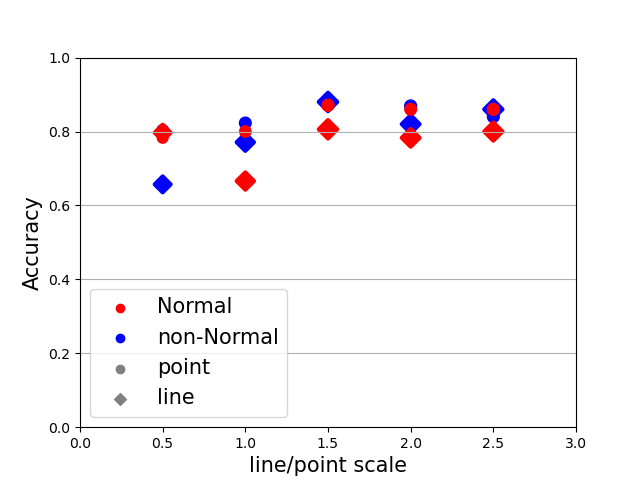}}
		\subfloat[]{\includegraphics[width=0.25\textwidth]{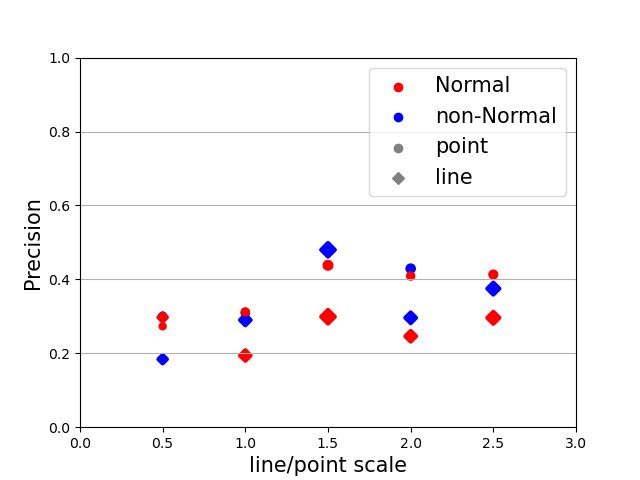}}
		\subfloat[]{\includegraphics[width=0.25\textwidth]{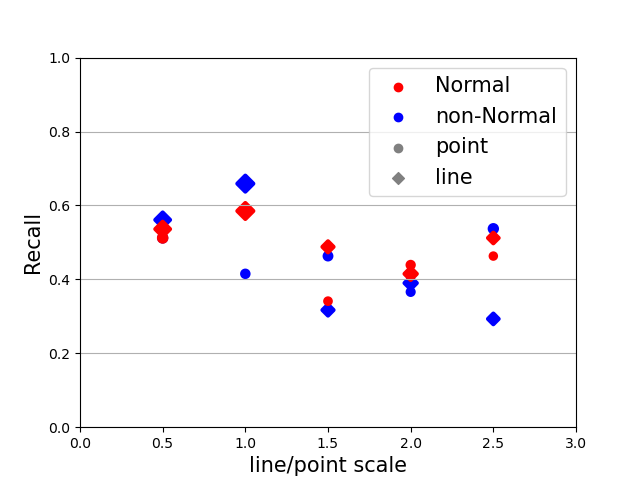}}
		\subfloat[]{\includegraphics[width=0.25\textwidth]{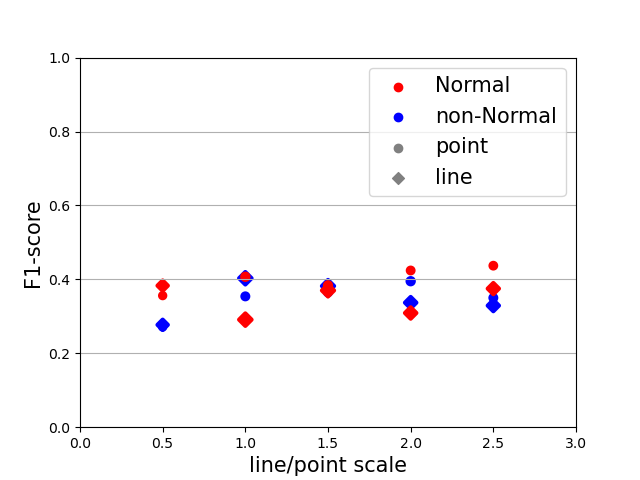}}
		\caption{The effects of curve scale, type and normalization operation on training process and performance. (a)(b)(c)(d) show the performance of different parameters in S2I on Flights.}
		\label{fig6}
	\end{figure*}
	
	\subsection{Experimental Setup}
	\textbf{Evaluation Metrics.} Based on the wide range of evaluation metrics used in object classification, the performance is usually reflected in four metrics, namely, accuracy, precision, recall, and F1-score. 
	
	These metrics are calculated using a confusion matrix, which is defined in Table \ref{table1}.
	
	\renewcommand{\arraystretch}{1.5} 
	\begin{table}[t]
		\centering
		\fontsize{8}{10}\selectfont
		\caption{Confusion Matrix}
		\begin{tabular}{c|cc}
			\Xhline{1.2pt}
			\diagbox{AC}{PC}& Positive & Negative \\ 
			\Xhline{1.2pt}
			True & True positive (TP) & False negative (FN) \\
			False & False positive (FP) & True negative (TN) \\
			\Xhline{1.2pt}
		\end{tabular}
		\label{table1}
	\end{table}

	The AC and PC in Table \ref{table1} indicate Actual Condition and Predicted Condition. 
	TP and TN are the test results that accurately predict positive and negative samples. FN and FP denote the test results that inaccurately predict the positive and negative samples correspondingly.
	The relevant indicator is thus given by equation \ref{APRF}.
	
	\begin{equation}
		\label{APRF}
		\begin{aligned}
			Precison &= \frac{TP}{TP + FP}, \\
			Recall &= \frac{TP}{TP + FN}, \\
			Acc &= \frac{TP + TN}{TP + TN + FP + FN}, \\
			F1-score &= 2 \times \frac{Precision \times Recall}{Precision + Recall}.
		\end{aligned}
	\end{equation}

	\textbf{Datasets.} 
	Flights dataset is selected by this paper, which collects a large amount of ultra-long temporal flight test data.
	It contains two classes, namely, minority (positive) class and majority (negative) class, which consists of 700 samples (350 training and 350 testing samples). 
	There are 310 negative samples and 40 positive samples in the training set, and 309 negative and 41 positive samples in the test set. 
	Among all the samples, the minimum, maximum and average sample lengths are $ 24,628 $ , $477,888$ and $254,076$, respectively.
	
	After that, three datasets in UCR Archive \cite{dau2019ucr} are selected to ensure the GTDA's efficiency, including Earthquakes, HandOutlines, and Herring. The detail in these datasets is listed in Table \ref{table2} .
	
	\renewcommand{\arraystretch}{1.5} 
	\begin{table}[t]
		
		\centering
		\fontsize{7.5}{10}\selectfont
		\begin{threeparttable}
			\caption{Statistics of the three temporal datasets and the Flights dataset}
			\label{table2}
			\label{tab:perfor}
			\begin{tabular}{c|ccccc}
				\Xhline{1.2pt}
				\multirow{1}{*}{Name}&Type&Length&Train&Test&Class\\
				\Xhline{1.2pt}
				
				\multirow{1}{*}{Earthquakes}& Sensor &$512$&$322$&$139$&$2$\\
				\multirow{1}{*}{HandOutlines}& Image &$2,709$&$1,000$&$370$&$2$\\
				\multirow{1}{*}{Herring}& Image &$512$&$64$&$64$&$2$\\
				\hline
				\multirow{1}{*}{\textbf{Flights}}& Sensor &$254,076$&$350$&$350$&$2$\\
				\Xhline{1.2pt}
			\end{tabular}
			\
		\end{threeparttable}

	\end{table}
	
	\textbf{Implementation Details.} 
	ResNet-34 \cite{he2016deep} is chosen as the backbone for this task because the network architecture was not the focus of our study and, on the other hand, ResNet-34 \cite{he2016deep} is used by many classification networks. 
	In addition, we set the same hyperparameters for all experiments. Specifically, the learning rate was set to 0.0002 and the epoch's number to 50. The relevant parameters for CRD, S2I and VBL will be elaborated in the ablation experiments section.

	\renewcommand{\arraystretch}{1.5} 
	\begin{table*}[tp]
		\centering
		\fontsize{7.5}{10}\selectfont
		\begin{threeparttable}
			\caption{The performance of typical classification methods and the proposed GTDA on the four datasets. All experiments use S2I to convert temporal data to images.}
			\label{experiment}
			\begin{tabular}{c|c | ccc | cccc | cccc }
				\Xhline{1.2pt}
				\multirow{2}{*}{Method}&\multirow{2}{*}{Backbone}&\multirow{2}{*}{S2I}&\multirow{2}{*}{CRD}&\multirow{2}{*}{VBL}
				&\multicolumn{4}{c|}{Earthquakes} & \multicolumn{4}{c}{Herring}  \\
				\cline{6-13}
				&&&&&  Acc&Precision&Recall&F1-score&  Acc&Precision&Recall&F1-score \\
				\Xhline{1.2pt}
				VGG-16 \cite{simonyan2014very}   & VGG-16    & \ding{51}& \ding{55}& \ding{55}&\color{red}{\textbf{0.705}}&\textbf{0.333}&0.171& 0.226&\multicolumn{4}{c}{Does not converge} \\
				\rowcolor{lightgray}	GTDA  & VGG-16    & \ding{51}& \ding{51}& \ding{51}&0.647& 0.325&\textbf{0.371}&\textbf{0.347}&\multicolumn{4}{c}{Does not converge} \\
				\hline
				ResNet-34 \cite{he2016deep} & ResNet-34 & \ding{51}& \ding{55}& \ding{55}&0.619& 0.304&0.400& 0.346&0.625& 0.533&0.615&0.571 \\
				\rowcolor{lightgray}			GTDA	  & ResNet-34 & \ding{51}& \ding{51}& \ding{51}&\textbf{0.676}&\textbf{ 0.368}&\textbf{0.400}&\textbf{ 0.384}&\textbf{0.641}& \textbf{0.552}&\textbf{0.615}&\textbf{0.582} \\

				\Xhline{2.0pt}
				
				\multirow{2}{*}{Method}&\multirow{2}{*}{Backbone}&\multirow{2}{*}{S2I}&\multirow{2}{*}{CRD}&\multirow{2}{*}{VBL}
				&\multicolumn{4}{c|}{HandOutlines} & \multicolumn{4}{c}{Flights}  \\
				\cline{6-13}
				&&&&&  Acc&Precision&Recall&F1-score&  Acc&Precision&Recall&F1-score \\
				\Xhline{1.2pt}
				VGG-16 \cite{simonyan2014very}    & VGG-16    & \ding{51}& \ding{55}& \ding{55}&0.930&\textbf{0.953}&0.937& 0.945&\multicolumn{4}{c}{Does not converge} \\
				\rowcolor{lightgray}			GTDA	  & VGG-16    & \ding{51}& \ding{51}& \ding{51}&\textbf{0.943}&0.939&\textbf{0.974}&\textbf{0.956}&\textbf{0.831}& \textbf{0.333}& \textbf{0.439}&\textbf{0.379} \\
				\hline
				ResNet-34 \cite{he2016deep} & ResNet-34 & \ding{51}& \ding{55}& \ding{55}&0.935&\textbf{0.961}&0.936& 0.948&\color{red}{\textbf{0.863}}&0.267& 0.098& 0.143 \\
				\rowcolor{lightgray}		GTDA	  & ResNet-34 & \ding{51}& \ding{51}& \ding{51}&\textbf{0.949}&0.954&\textbf{0.966}&\textbf{0.960}&0.857&\textbf{0.404}& \textbf{0.463}&\textbf{0.432} \\

				\Xhline{1.2pt}
			\end{tabular}
		\end{threeparttable}
		
	\end{table*}

	\subsection{Ablation Study}
	
	We perform ablation experiments in two ways. First, ignoring resampling as well as reweighting, that is, not considering CRD and VBL, we verify the effects of different parameters on S2I and then further compare the effects of S2I on the experiments. Second, on the basis of the S2I, we use S2I+ResNet34\cite{he2016deep} to fully explore the effects of CRD and VBL on model performance.
	
	\textbf{The Influence of S2I Parameters.} The scale and type of curve, and whether or not the data is normalized are all taken into account. Specifically, we use the pyplot function in Matplotlib to set the width of of the sensor curve in the image, with the widths being set to $ \{ 0.5, 1.0, 1.5, 2.0, 2.5 \} $. The type of curve includes line or point. Similarly, the point size corresponds to $ \{ 0.5, 1.0, 1.5, 2.0, 2.5 \} $, and the point uses the parameter names markersize in pyplot. The data is normalized using Min-Max Normalization, which scales the range of values for each feature point to be normalized to $ [0,1] $. For each sample, the value of the $ k^{\text{th}} $ feature point is given by the following equation:
	
	\begin{equation}
		\label{Normalization}
		\begin{aligned}
			\hat{x}_{ik} = \frac{x_{ik} - min \{ x_i \} }{max \{ x_i \} - min \{ x_i \}},
		\end{aligned}
	\end{equation}
	
	\noindent where $ max\{ x_i \} $ and $ min\{ x_i \} $ denote the maximum and minimum values of all feature points in the $ i^{\text{th}} $ sample, respectively. Equation \ref{Normalization} allows $ x_{ik} $ to be mapped to $ [0,1] $. 
	
	We denote the type of curve as the set $ \mathbb{A} $, where $ \mathbb{A} = \{ 0.5, 1.0, 1.5, 2.0, 2.5 \} $. Similarly, the type $ \mathbb{B} = \{ \text{line}, \text{point} \} $ and the normalization operation $ \mathbb{C} = \{ \text{Normal}, \text{non-Normal} \} $ are defined. 
	Thus, all operations of S2I can be given by formula \ref{Set}.
	
	\begin{equation}
		\label{Set}
		\begin{aligned}
			\mathbb{A} \times \mathbb{B} \times \mathbb{C} = \{ (x, y, z) | x \in \mathbb{A}, y \in \mathbb{B}, z \in \mathbb{C} \} ,
		\end{aligned}
	\end{equation}
	
	\noindent where $ \times $ denotes Cartesian product. 
	The optional parameter details are listed in Fig. \ref{fig5} (a).
	Further, Fig. \ref{fig5} (b), (c), and (d) are drawn to more intuitively distinguish between the different operations. More specifically, $ (0.5, \text{line}, \text{Normal}) $ is set in the Fig. \ref{fig5} (b). The parameters of Fig. \ref{fig5} (c) and (d) are set to $ (1.5, \text{point}, \text{Normal}) $ and $ (2.5, \text{line}, \text{non-Normal}) $, respectively.

	As shown in Fig. \ref{fig6}, we conduct experiments with different parameters and use diverse metric values in formula \ref{APRF}. 
	The vertical axis presents the metric values on the Flights dataset. The horizontal vertical presents different scale of curve, which is generated by temporal data after S2I. In the same way, shape and color of the icon correspond to curve type and normalization operation in data, respectively. Considering the four evaluation metrics, when $ (x,y,z) $ is equal to $ (1.5, line, non-Normal) $, it has a better effect. So $ (1.5, line, non-Normal) $ is selected as the setting of S2I. 
	
	\renewcommand{\arraystretch}{1.5} 
	\begin{table}[tp]
		\centering
		\fontsize{7.5}{10}\selectfont
		\begin{threeparttable}
			\caption{The effectiveness of the CRD and VBL on Flights dataset}
			\label{CRDandVBL}
			\begin{tabular}{cc |  cccc}
				\Xhline{1.2pt}
				CRD&VBL&Acc&Precision&Recall&F1-score\\
				\Xhline{0.5pt}
				\ding{55} & \ding{55} &0.863&0.267&0.098&0.143\\
				\ding{51} & \ding{55} &\textbf{0.880}&\textbf{0.480}&0.317&0.382\\
				\ding{55} &\ding{51}  &0.711&0.254&\textbf{0.756}&0.380\\
				\ding{51} &\ding{51}  &0.857&0.404&0.463&\textbf{0.432}\\
				\Xhline{1.2pt}
			\end{tabular}
		\end{threeparttable}
		
	\end{table}
	
	\textbf{CRD and VBL analysis.} 
	Table \ref{CRDandVBL} describes the detection results of each component with the S2I module using ResNet-34 as the backbone.
	For a fair comparison, we trained the backbone using the standard Softmax Cross-Entropy loss as well as without using the resample method. First, we verify the performance of CRD. CRD improves the performance of all evaluation indicators comparing to the baseline, i.e., +0.017 for accuracy, +0.213 for precision, +0.219 for recall and +0.239 for F1-score, respectively. 
	The above results demonstrate that CRD can significantly adjust the decision boundary and help the network train the minority samples better.
	We then investigate the efficacy of VBL.
	The recall and F1-score increased by 0.658 and 0.237 compared to the baseline.
	Analyzing the above comparative data, CRD improves performance from the precision perspective, while VBL concentrates more on the recall of abnormal samples.
	Next, we verify the persuasiveness of the combination of CRD and VBL based on S2I and ResNet34. 
	Specifically, CRD and VBL work collaboratively and improve 0.137 for precision, 0.365 for recall and 0.289 for F1-score comparing to the baseline. 
	Notably, the combination of CRD and VBL increase 0.150 for precision comparing to the VBL only, and 0.146 for recall comparing to the CRD only. Therefore, the complete method achieve the best of F1-score, i.e., +0.289 to the baseline, +0.05 to the CRD, and +0.052 to the VBL.

	\begin{figure*}[t]
		
		\centering                     
		\includegraphics[width=0.8\textwidth]{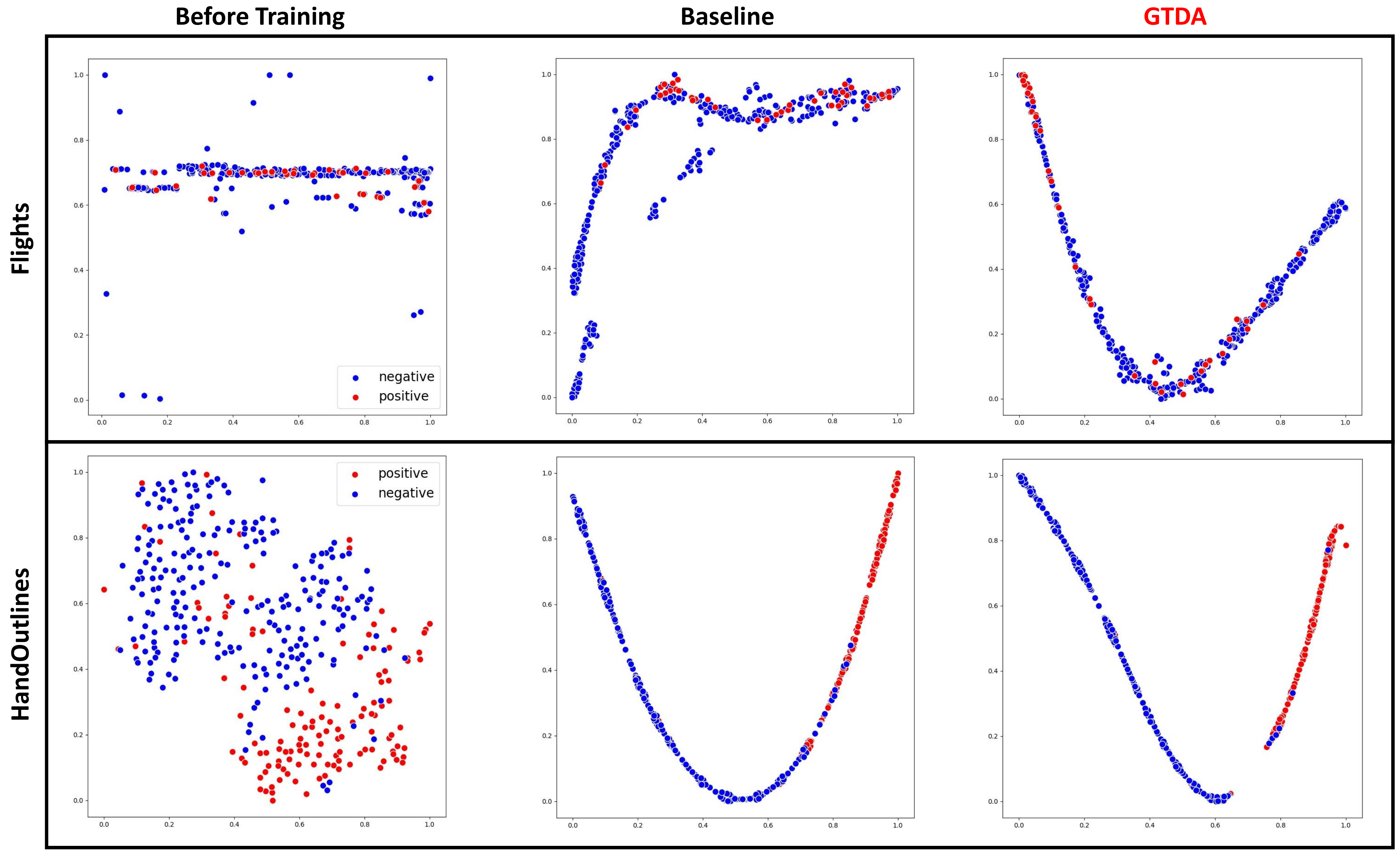}
		
		\caption{The visualization of Flights and HandOutlines datasets with t-SNE, and results of the baseline and the proposed GTDA in these datasets.}
		\label{fig7}
	\end{figure*}

	\subsection{Experimental Results}
	In this section, we test the performance of CRD and VBL on Flights and three other datasets with ResNet-34 and VGG-16 as baselines. 
	For a fair comparison, all the temporal samples of the four datasets is converted to image data via S2I.
	Table \ref{experiment} lists the statistical results of the four evaluation indicators. 
	The proposed GTDA indicates the CRD and VBL based on the baseline.
	Compare with baseline only, our method performs better. 
	Taking the F1-score as an example, the proposed module increases by 0.038, 0.121, 0.012, 0.009, 0.011, and 0.289 compared with the corresponding backbone and dataset, respectively.

	\subsection{Experimental Analysis}
	\textbf{ Qualitative Results.} For extremely imbalanced datasets, the accuracy cannot effectively evaluate the quality of the model fitting ability. Taking Flights dataset as an example, the 350 test set samples contain 309 negative samples and 41 positive samples. The imbalance ratio reaches $ 7.54:1 $. 
	It means that the accuracy rate is 0.883 when all samples are predicted as negative by the model.
	This is an inaccurate evaluation of the model in an imbalanced dataset. 
	F1-score takes into account the precision and recall, and the evaluation is fair and effective. 
	The two bolded values in red in Table \ref{experiment} indicate that the backbone has a higher accuracy without CRD and VBL. 
	But this accuracy cannot express the degree of model fit, because they all have lower F1-score. 
	That is, both models predict the positive samples as negative samples more.
	Due to the insufficient number of layers of $ \text{VGGNet-16 \cite{simonyan2014very}} $ itself and the weak model fitting ability, it cannot effectively fit samples of different classes in the Herring dataset, and all samples are predicted to be the majority class. 
	In Flights dataset, $ \text{VGGNet-16 \cite{simonyan2014very}} $ still fails to fit, but with the help of the proposed modules (CRD+VBL), it fits efficiently and performs well. Specifically, ResNet-34 has better fitting ability than $ \text{VGGNet-16 \cite{simonyan2014very}} $ objectively, but in F1-score, $ \text{ResNet-34} $ \cite{he2016deep} is 0.236 lower than GTDA.
	
	\textbf{Embedding Visualization.}
	As show in Fig. \ref{fig7}, t-SNE \cite{van2008visualizing} is used to visualize the performance of the proposed GTDA on Flights and HandOutlines datasets. 
	For efficient comparison, we first visualize all samples in the initial dataset. 
	The before training column in Fig. \ref{fig7} shows that the Handoutlines dataset has more obvious data separability than Flights after t-SNE dimensionality reduction.
	Flights dataset is almost inseparable after dimensionality reduction. 
	Then, the baseline is used after S2I to train the dataset.
	Finally, we added CRD and VBL to the baseline. 
	And we visualize these results. 
	In Flights dataset, the baseline mixes most of the positive samples with some negative samples, which is more obvious in the upper part of the figure. 
	And the proposed GTDA concentrates more positive samples in the upper left of the figure. Combined with Table \ref{experiment}, GTDA has a higher F1-score than the baseline. 
	In the HandOutlines dataset, the model trained by GTDA clearly separates positive instances from negative after visualization, but the baseline does not.

	\section{Conclusion}

	This paper focuses on anomaly detection for ultra-long temporal flight test data.
	To this end, we develop a general framework (GTDA) for studying time-series data using CV-based methods. 
	It contains three modules: imaging temporal data module (S2I), resampling module (CRD) and reweighting module (VBL). 
	Specifically, S2I converts temporal data into images. 
	CRD oversamples minority class by clustering to adjust the decision boundary coarsely. 
	VBL adaptively attaches different weights for each class to adjust the effect intensity on the decision boundary.
	Extensive experiments demonstrate the GTDA's eminence in ultra-long temporal data anomaly detection.
	Besides, CRD and VBL promote the precision and recall of the model, respectively. The synergy of CRD and VBL can improve the F1-score.
	In future work, we will concentrate on exploring the role of cross-modal flight test data in anomaly detection, and further investigate how to extract effective features from multi-modal flight test data.

	\bibliographystyle{IEEEtran}
	\bibliography{reference.bib}

\end{document}